\documentclass[11pt,a4paper]{article}
\usepackage[hyperref]{acl2019}
\usepackage{times}
\usepackage{latexsym}
\usepackage{booktabs}
\usepackage{graphicx} 
\usepackage{url}
\usepackage{enumitem}
\usepackage{amsmath}
\aclfinalcopy 
\def\aclpaperid{1938} 

\title{Deep Unknown Intent Detection with Margin Loss}

\author{Ting-En Lin, Hua Xu \\
  State Key Laboratory of Intelligent Technology and Systems, \\
 Department of Computer Science and Technology, Tsinghua University, Beijing, China \\
  Beijing National Research Center for Information Science and Technology \\
  {\tt lte17@mails.tsinghua.edu.cn, xuhua@tsinghua.edu.cn} \\ }

\date{}

\begin{document}
\maketitle
\begin{abstract}
Identifying the unknown (novel) user intents that have never appeared in the training set is a challenging task in the dialogue system. In this paper, we present a two-stage method for detecting unknown intents. We use bidirectional long short-term memory (BiLSTM) network with the margin loss as the feature extractor. With margin loss, we can learn discriminative deep features by forcing the network to maximize inter-class variance and to minimize intra-class variance. Then, we feed the feature vectors to the density-based novelty detection algorithm, local outlier factor (LOF), to detect unknown intents. Experiments on two benchmark datasets show that our method can yield consistent improvements compared with the baseline methods.
\end{abstract}

\section{Introduction}
In the dialogue system, it is essential to identify the unknown intents that have never appeared in the training set. We can use those unknown intents to discover potential business opportunities. Besides, it can provide guidance for developers and accelerate the system development process. However, it is also a challenging task. On the one hand, it is often difficult to obtain prior knowledge about unknown intents due to lack of examples. On the other hand, it is hard to estimate the exact number of unknown intents. In addition, since user intents are strongly guided by prior knowledge and context, modeling high-level semantic concepts of intent is still problematic.

Few previous studies are related to unknown intents detection. For example, \citet{Kim2018JointLO} try to optimize the intent classifier and out-of-domain detector jointly, but out-of-domain samples are still needed. The generative method \cite{Yu2017OpenCategoryCB} try to generate positive and negative examples from known classes by using adversarial learning to augment training data. However, the method does not work well in the discrete data space like text, and a recent study \cite{2018arXiv181009136N} suggests that this approach may not work well on real-world data. \citet{Brychcin2017UnsupervisedDA} try to model intents through clustering. Still, it does not make good use of prior knowledge provided by known intents, and clustering results are usually unsatisfactory.

Although there is a lack of prior knowledge about unknown intents, we can still leverage the advantage of known label information. \citet{Scheirer2013TowardOS,Fei2016BreakingTC} suggest that a \emph{m}-class classifier should be able to reject examples from unknown class while performing \emph{m}-class classification tasks. The reason is that not all test classes have appeared in the training set, which forms a (\emph{m}+1)-class classification problem where the (\emph{m}+1)$^{th}$ class represents the unknown class. This task is called open-world classification problem. The main idea is that if an example dissimilar to any of known intents, it is considered as the unknown. In this case, we use known intents as prior knowledge to detect unknown intents and simplify the problem by grouping unknown intents into a single class.

\citet{Bendale2016TowardsOS} further extend the idea to deep neural networks (DNNs). \citet{Shu2017DOCDO} achieve the state-of-the-art performance by replacing the softmax layer of convolution neural network (CNN) with a 1-vs-rest layer consist of sigmoid and tightening the decision threshold of probability output for detection.
\begin{figure*}[t]
  \centering
  \includegraphics[width=0.999\linewidth]{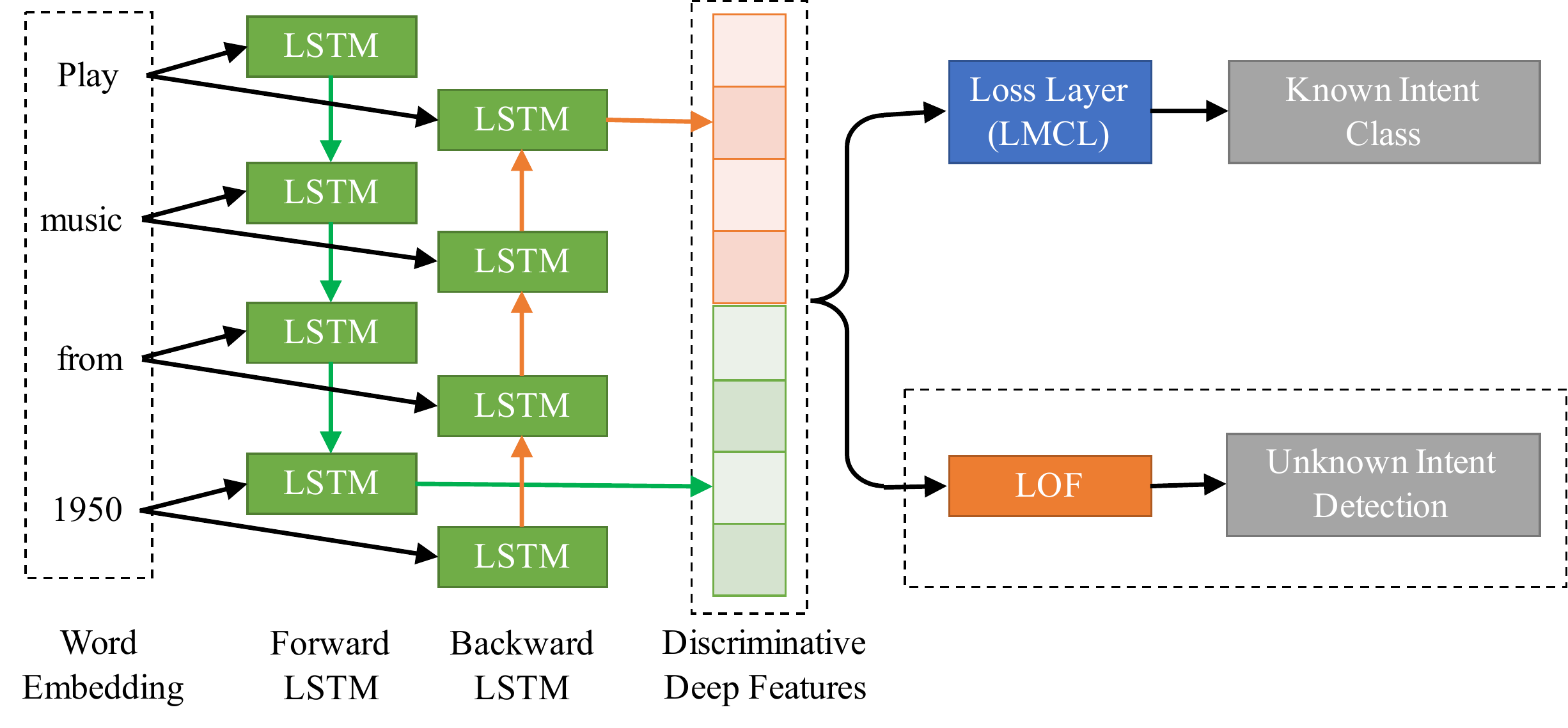}
  \caption{The architecture of the proposed two-stage method. We acquire intent representation by training an intent classifier on known intent with BiLSTM and learn discriminative deep features through LMCL. Then, we use LOF to detect unknown intents during the testing stage. \label{model}}
\end{figure*}

DNN such as BiLSTM \cite{Goo2018SlotGatedMF, Wang2018ABB} has demonstrated the ability to learn high-level semantic features of intents. Nevertheless, it is still challenging to detect unknown intents when they are semantically similar to known intents. The reason is that softmax loss only focuses on whether the sample is correctly classified, and does not require intra-class compactness and inter-class separation. Therefore, we replace softmax loss with margin loss to learn more discriminative deep features. 

The approach is widely used in face recognition \cite{DBLP:conf/icml/LiuWYY16, DBLP:conf/cvpr/LiuWYLRS17, DBLP:journals/corr/RanjanCC17}. It forces the model to not only classify correctly but also maximize inter-class variance and minimize intra-class variance. Concretely, we use large margin cosine loss (LMCL) \cite{DBLP:conf/cvpr/WangWZJGZL018} to accomplish it. It formulates the softmax loss into cosine loss with $L_2$ norm and further maximizes the decision margin in the angular space. Finally, we feed the discriminative deep features to a density-based novelty detection algorithm, local outlier factor (LOF), to detect unknown intents.

We summarize the contributions of this paper as follows. First, we propose a two-stage method for unknown intent detection with BiLSTM. Second, we introduce margin loss on BiLSTM to learn discriminative deep features, which is suitable for the detection task. Finally, experiments conducted on two benchmark dialogue datasets show the effectiveness of the proposed method. 

\section{Proposed Method}
\subsection{BiLSTM}
To begin with, we use BiLSTM \cite{mesnil2015using} to train the intent classifier and use it as feature extractor. Figure \ref{model} shows the architecture of the proposed method. Given an utterance with maximum word sequence length $\ell $, we transform a sequence of input words $w_{1:\ell}$ into m-dimensional word embedding $v_{1:\ell}$, which is used by forward and backward LSTM to produce feature representations $x$: 
\begin{align}
\overrightarrow{x_t} &= LSTM(v_t,\overrightarrow{c_{t-1}}), \nonumber \\
\overleftarrow{x_t} &= LSTM(v_t,\overleftarrow{c_{t+1}}), \nonumber \\
x &=[\overrightarrow{x_{\ell}};\overleftarrow{x_{1}}], 
\end{align} 
where $v_{t}$ denotes the word embedding of input at time step $t$. $\overrightarrow{x_{t}}$ and $\overleftarrow{x_{t}}$ are the output vector of forward and backward LSTM respectively.  $\overrightarrow{c_{t}}$ and $\overleftarrow{c_{t}}$ are the cell state vector of forward and backward LSTM respectively.

We concatenate the last output vector of forward LSTM ${\overrightarrow{x_{\ell}}}$ and the first output vector of backward LSTM $\overleftarrow{x_{1}}$ into $x$ as the sentence representation. It captures high-level semantic concepts learned by the model. We take $x$ as the input of the next stage. 

\begin{table*}[t!]
\centering
\begin{tabular}{llllllll}
\toprule
  Dataset & Classes & Vocabulary & \#Training & \#Validation & \#Test & Class distribution \\
  \hline
  SNIPS & 7 & 11,971 & 13,084 & 700 & 700 & Balanced \\
  ATIS & 18 & 938 & 4,978 & 500 & 893 & Imbalanced\\
\bottomrule
\end{tabular}
\caption{ \label{data-stat-table}  Statistics of SNIPS and ATIS dataset. \# indicates the total number of utterances.}
\end{table*}

\subsection{Large Margin Cosine Loss (LMCL)}
At the same time, we replace the softmax loss of BiLSTM with LMCL \cite{2018arXiv181009136N}. We define LMCL as the following:
\begin{align}
&\mathcal{L}_{LMC} = \nonumber \\ 
&\frac{1}{N} \sum_{i} - \log \frac{e^{s \cdot  ({\cos{(\theta_{y_i,i}})}-m)}}{e^{s \cdot ({\cos{(\theta_{y_i,i}})-m})} + \sum_{j \neq y_i} e^{s \cdot \cos{\theta_{j,i}}}},
\end{align}
constrained by  
\begin{align}
 \cos(\theta_j, i) &= W_j^Tx_i , \nonumber \\ 
   W = \frac{W^\star}{||W^\star||}, & \quad  x = \frac{x^\star}{||x^\star||}, 
\end{align}
where N denotes the number of training samples, $y_i$ is the ground-truth class of the $i$-th sample, $s$ is the scaling factor, $m$ is the cosine margin, $W_j$ is the weight vector of the $j$-th class, and $\theta_j$ is the angle between $W_j$ and $x_i$.

LMCL transforms softmax loss into cosine loss by applying L2 normalization on both features and weight vectors. It further maximizes the decision margin in the angular space. With normalization and cosine margin, LMCL forces the model to maximize inter-class variance and to minimize intra-class variance. Then, we use the model as the feature extractor to produce discriminative intent representations.

\subsection{Local Outlier Factor (LOF)}
Finally, because the discovery of unknown intents is closely related to the context, we feed discriminative deep features $x$ to LOF algorithm \cite{breunig2000lof} to help us detect unknown intents in the context with local density. We compute LOF as the following:
\begin{align}
    \text{LOF}_k(A) = \frac{\sum_{B \in N_k(A)} \frac{\text{lrd}(B)}{\text{lrd}(A)}}{|N_k(A)|},
\end{align}
where $N_k(A)$ denotes the set of k-nearest neighbors and \emph{lrd} denotes the local reachability density. We define \emph{lrd} as the following:
\begin{align}
    \text{lrd}_k(A) = \frac{|N_{k}(A)|}{\sum_{B\in N_k(A)}} \text{reachdist}_k(A,B),
\end{align}
where $\text{lrd}_k(A)$ denotes the inverse of the average reachability distance between object A and its neighbors. We define $\text{reachdist}_k(A,B)$ as the following:
\begin{align}
    \text{reachdist}_k(A,B) = \max\{{\text{k-dist}(B), d(A,B)}\},
\end{align}
where d(A,B) denotes the distance between A and B, and k-dist denotes the distance of the object A to the $k^{th}$ nearest neighbor. If an example's local density is significantly lower than its k-nearest neighbor's, it is more likely to be considered as the unknown intents. 
\begin{table*}[t!]
\centering
\begin{tabular}{lllllllllll}
   & SNIPS &   &  & ATIS &   &  & \\
\toprule
\% of known intents & 25\% & 50\% & 75\%  & 25\% & 50\% & 75\%  \\
  \hline
MSP & 0.0 & 6.2 & 8.3 & 8.1 & 15.3 & 17.2 \\
DOC & 72.5 & 67.9 & 63.9 & 61.6 & 62.8 & 37.7 \\
DOC (Softmax) & 72.8 & 65.7 & 61.8 & 63.6 & 63.3 & 38.7 \\
LOF (Softmax) & 76.0 & 69.4 & 65.8 & 67.3 & 61.8 & 38.9 \\
LOF (LMCL) & \textbf{79.2} & \textbf{84.1} & \textbf{78.8} & \textbf{69.6} & \textbf{63.4} & \textbf{39.6} \\
\bottomrule
\end{tabular}
\caption{ \label{result-table}  Macro f1-score of unknown intent detection with different proportion (25\%, 50\% and 75\%) of classes are treated as known intents on SNIPS and ATIS dataset.}
\end{table*}

\begin{figure*}[t]
  \centering
  \includegraphics[width= 0.999 \linewidth]{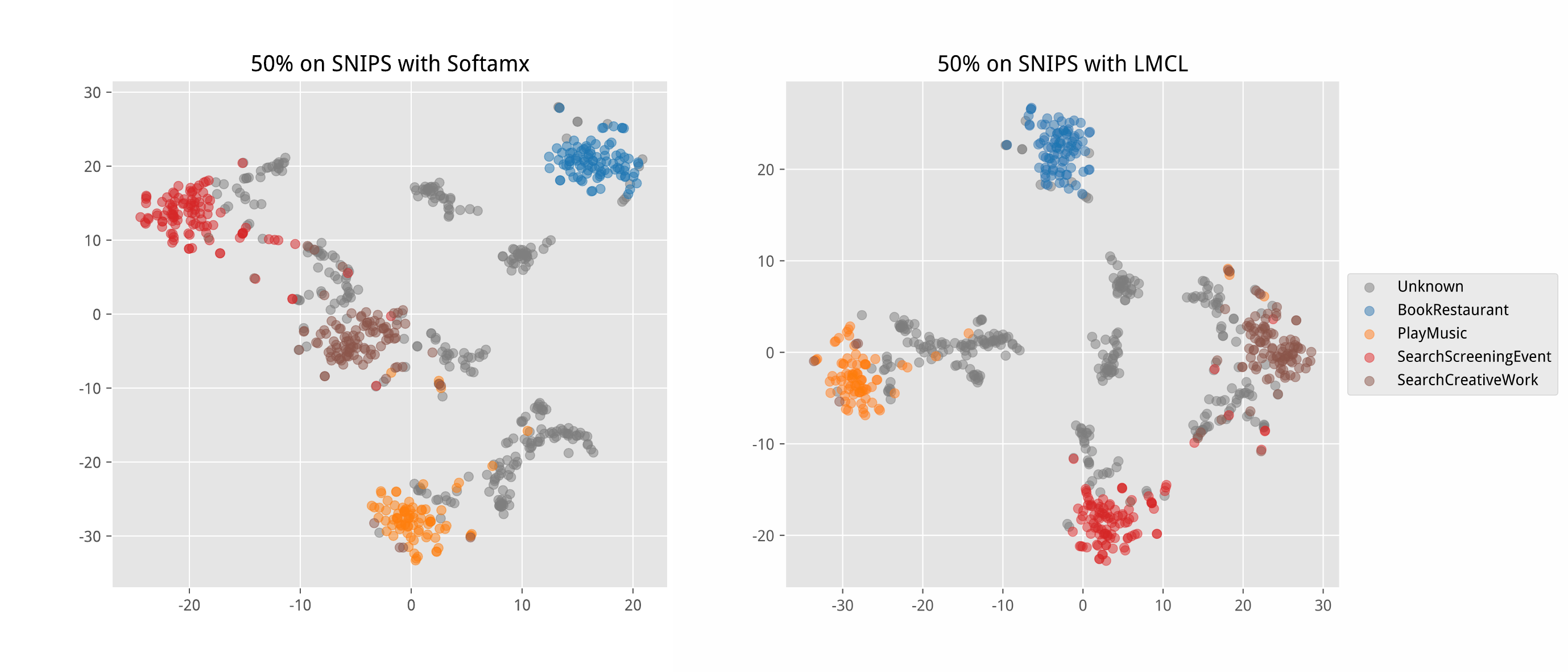}
\caption{ \label{mat} Visualization of deep features  learned with softmax and LMCL on SNIPS dataset.}
\end{figure*}
                                                                                                                                                                                                                                                                                                                                                                                                                                                                                                                                                                                                                                              
\section{Experiments}
\subsection{Datasets}
We have conducted experiments on two publicly available benchmark dialogue datasets, including SNIPS and ATIS \cite{Tr2010WhatIL}. The detailed statistics are shown in Table~\ref{data-stat-table}. 

\textbf{SNIPS \footnote{https://github.com/snipsco/nlu-benchmark/tree/master/2017-06-custom-intent-engines}}
SNIPS is a personal voice assistant dataset which contains 7 types of user intents across different domains.

\textbf{ATIS (Airline Travel Information System) \footnote{https://github.com/yvchen/JointSLU/tree/master/data}} ATIS dataset contains recordings of people making reservations with 18 types of user intent in the flight domain.

\subsection{Baselines}
We compare our methods with state-of-the-art methods and a variant of the proposed method.

\begin{enumerate}[noitemsep,topsep=0pt]
  \item \textbf{Maximum Softmax Probability (MSP)}  \cite{hendrycks2016baseline} Consider the maximum softmax probability of a sample as the score, if a sample does not belong to any known intents, its score will be lower. We calculate and apply a confidence threshold on the score as the simplest baseline where the threshold is set as 0.5. 
  \item \textbf{DOC} \cite{Shu2017DOCDO} 
It is the state-of-the-art method in the field of open-world classification. It replaces softmax with sigmoid activation function as the final layer. It further tightens the decision boundary of the sigmoid function by calculating the confidence threshold for each class through statistics approach.
  \item \textbf{DOC (Softmax)} A variant of DOC. It replaces the sigmoid activation function with softmax. 
  \item \textbf{LOF (Softmax)} A variant of the proposed method for ablation study. We use softmax loss to train the feature extractor rather than LMCL.
\end{enumerate}

\subsection{Experimental Settings}
We follow the validation setting in \cite{Fei2016BreakingTC, Shu2017DOCDO} by keeping some classes in training as unknown and integrate them back during testing. Then we vary the number of known classes in training set in the range of 25\%, 50\%, and 75\% classes and use all classes for testing. 

To conduct a fair evaluation for the imbalanced dataset, we randomly select known classes by weighted random sampling without replacement in the training set. If a class has more examples, it is more likely to be chosen as the known class. Meanwhile, the class with fewer examples still have a chance to be selected. Other classes are regarded as unknown and we will remove them in the training and validation set.

We initialize the embedding layer through GloVe \cite{pennington2014glove} pre-trained word vectors \footnote{http://nlp.stanford.edu/projects/glove/}. For BiLSTM model, we set the output dimension as 128 and the maximum epoch as 200 with early stop. For LMCL and LOF, we follow the original setting in their paper. We use macro f1-score as the evaluation metric and report the average result over 10 runs. We set the scaling factor $s$ as 30 and cosine margin $m$ as 0.35, which is recommended by \citet{wang2018additive}.

 \subsection{Results and Discussion}
We show the experiment results in Table \ref{result-table}. Firstly, our method consistently performs better than all baselines in all settings. Compared with DOC, our method improves the macro f1-score on SNIPS by 6.7\%, 16.2\% and 14.9\% in 25\%, 50\%, and 75\% setting respectively. It confirms the effectiveness of our two-stage approach. 

Secondly, our method is also better than LOF (Softmax). In Figure \ref{mat}, we use t-SNE \cite{maaten2008visualizing} to visualize deep features learned with softmax and LMCL. We can see that the deep features learned with LMCL are intra-class compact and inter-class separable, which is beneficial for novelty detection algorithms based on local density.

Thirdly, we observe that on the ATIS dataset, the performance of unknown intent detection dramatically drops as the known intent increases. We think the reason is that the intents of ATIS are all in the same domain and they are very similar in semantics (e.g., flight and  flight\_no). The semantics of the unknown intents can easily overlap with the known intents, which leads to the poor performance of all methods.

Finally, compared with ATIS, our approach improve even better on SNIPS. Since the intent of SNIPS is originated from different domains, it causes the DNN to learn a simple decision function when the known intents are dissimilar to each other. By replacing the softmax loss with the margin loss, we can push the network to further reduce the intra-class variance and the inter-class variance, thus improving the robustness of the feature extractor.

\section{Conclusion}
In this paper, we proposed a two-stage method for unknown intent detection. Firstly, we train a BiLSTM classifier as the feature extractor. Secondly, we replace softmax loss with margin loss to learn discriminative deep features by forcing the network to maximize inter-class variance and to minimize intra-class variance. Finally, we detect unknown intents through the novelty detection algorithm. We also believe that broader families of anomaly detection algorithms are also applicable to our method. 

Extensive experiments conducted on two benchmark datasets show that our method can yield consistent improvements compared with the baseline methods.  In future work, we plan to design a solution that can identify the unknown intent from known intents and cluster the unknown intents in an end-to-end fashion.

\section*{Acknowledgments}
This paper is funded by National Natural Science Foundation of China (Grant No: 61673235) and National Key R\&D Program Projects of China (Grant No: 2018YFC1707600). We would like to thank the anonymous reviewers and Yingwai Shiu for their valuable feedback. 
\bibliography{acl2019}
\bibliographystyle{acl_natbib}

\end{document}